%% file: main.tex
\documentclass[11pt,a4paper]{article}

\usepackage[hyperref]{acl2019}
\usepackage{times}
\usepackage{latexsym}
\usepackage{nicefrac}

\aclfinalcopy 

\input{math_commands.tex} 
\input{tikz}

\usepackage[textsize=tiny]{todonotes}
\usepackage{url}
\usepackage{graphicx}
\usepackage{caption}
\usepackage{subcaption}
\usepackage{bm,dsfont}
\usepackage{amsmath}
\usepackage{units}
\usepackage[capitalize]{cleveref}
\creflabelformat{equation}{#2#1#3} 

\title{Transcoding compositionally: using attention to find more generalizable solutions}
\author{
  Kris Korrel \\
  University of Amsterdam \\
  {\tt kris.korrel@gmail.com}
  \\\And
    Dieuwke Hupkes \\
  University of Amsterdam \\
  {\tt d.hupkes@uva.nl}
  \\\AND
    Verna Dankers \\
  University of Amsterdam \\
  {\tt verna.dankers@gmail.com}
  \\\And
    Elia Bruni \\
  Universitat Pompeu Fabra \\
  {\tt elia.bruni@gmail.com}
  }
\date{}

\begin{document}
    \maketitle
    \input{abstract}
    \input{introduction}

    \input{related_work}

    \input{model}
    \input{lookup_tables}

    \input{scan}

    \input{conclusion}
    \section*{Acknowledgements}
    We are grateful to Kristina Gulordava for ideas and feedback.
DH is funded by the Netherlands Organization for Scientific Research (NWO),
through a Gravitation Grant 024.001.006 to the Language in Interaction Consortium. EB is funded by the European Union's Horizon 2020 research and innovation program under the Marie Sklodowska-Curie grant agreement No 790369 (MAGIC).
    \bibliographystyle{acl_natbib}
    \bibliography{refs}
\end{document}

%% file: math_commands.tex
\usepackage{amsmath,amsfonts,bm}

\def\eqref#1{equation~\ref{#1}}

\def\1{\bm{1}}

\DeclareMathAlphabet{\mathsfit}{\encodingdefault}{\sfdefault}{m}{sl}
\SetMathAlphabet{\mathsfit}{bold}{\encodingdefault}{\sfdefault}{bx}{n}



%% file: tikz.tex
\usepackage[mode=buildnew]{standalone}
\usepackage{tikz}
\usetikzlibrary{patterns}
\usepackage{pgfplots}
\pgfplotsset{
    discard if not/.style 2 args={
        x filter/.code={
            \edef\tempa{\thisrow{#1}}
            \edef\tempb{#2}
            \ifx\tempa\tempb
            \else
            \def\pgfmathresult{inf}
            \fi
        }
    },ymin=0,ymax=1,
    compat=newest
}
\usepgfplotslibrary{fillbetween}
\usepackage{pgfplotstable}
\pgfplotsset{compat=1.13}
\usepgfplotslibrary{colormaps} 
\definecolor{awesome}{rgb}{1.0, 0.13, 0.32}
\definecolor{magenta_mtplotlib}{rgb}{1.0, 0.0, 1.0}
\definecolor{green_mtplotlib}{rgb}{0.0, 0.5019607843137255, 0.0}
\definecolor{blue_mtplotlib}{rgb}{0.0, 0.0, 1.0}

%% file: abstract.tex
\begin{abstract}
While sequence-to-sequence models have shown remarkable generalization power across several natural language tasks, their construct of solutions are argued to be less compositional than human-like generalization.
In this paper, we present seq2attn, a new architecture that is specifically designed to exploit attention to find compositional patterns in the input.
In seq2attn, the two standard components of an encoder-decoder model are connected via a \textit{transcoder}, that modulates the information flow between them. 
We show that seq2attn can successfully generalize, without requiring any additional supervision, on two tasks which are specifically constructed to challenge the compositional skills of neural networks.
The solutions found by the model are highly interpretable, allowing easy analysis of both the types of solutions that are found and potential causes for mistakes.
We exploit this opportunity to introduce a new paradigm to test compositionality that studies the extent to which a model \textit{overgeneralizes} when confronted with exceptions.
We show that seq2attn exhibits such overgeneralization to a larger degree than a standard sequence-to-sequence model.
\end{abstract}

%% file: introduction.tex
\section{Introduction}
In recent years, deep artificial neural networks have been at the root of many successes in a wide variety of AI tasks,
including sequential tasks, for which encoder-decoder models are the de facto standard \citep{cho2014learning,sutskever2014sequence}.
These successes have also caused a renewed interest in the types of solutions that they learn \citep{bbnlp2018} and, in particular, have prompted the question: to what extent can their high accuracy be taken as evidence that they in fact understood the task they are modeling.
A number of recent studies argues that it cannot, when `understanding the task' is explained as understanding the implicit rules by which it is governed \citep[e.g., ][]{johnson2017clevr,lake2018generalization,livska2018memorize, feng2018pathologies, ravfogel2018can}.
More specifically, they argue that rather than understanding those implicit rules and being able to compositionally apply them, RNN models exploit biases in the data that are unrelated to the underlying system.
While the latter strategy is remarkably effective when large amounts of training data are available, the lack of understanding of the actual task leads to sample inefficiency, inability to transfer knowledge between tasks and difficulty to generalize to sequences that are drawn from the same rule space, but differ distributionally from the training data. 
Furthermore, the use of such strategies, which deviate largely from human approaches, that are typically compositional \citep{lake2015human}, makes it difficult to understand what a model does and when it may make a mistake.

In this work, we propose a new component that aims to address this particular weakness of seq2seq models. 
This component, which is a recurrent attention module that can be integrated in any form of encoder-decoder model, modulates the information flow from encoder to decoder. 
We test our module, which we dub \textit{seq2attn}, in a recurrent encoder-decoder model. 
Using two tasks that are designed such that their accuracy reflects directly whether the underlying rule-based system is learned -- the lookup table task \citep{livska2018memorize} and SCAN \citep{lake2018generalization,loula2018rearranging} -- we show that seq2attn strongly encourages rule-based behaviour, which is easily interpreted by studying the attention patterns generated by the module.
Additionally, we propose a new testing paradigm based on \textit{overgeneralization}, that can be used to gain more insights in the biases of a model which cannot be inferred from task success alone.

%% file: related_work.tex
\section{Related Work}
\label{sec:related_work}

\subsection{Compositional datasets}

The ability to learn and compositionally apply symbolic rules is considered to be an important prerequisite for understanding and modeling natural language. 
While (gated) recurrent neural networks are in principle capable of modeling compositional systems \citep[e.g.,][]{gers2001lstm,rodriguez2001simple}, whether they in fact do so when trained on large amounts of data to perform natural language processing tasks remains an open question.
Some positive results in this direction have been presented \citep[e.g.,][]{hupkes2018visualisation}, but a number of recent papers have argued that, rather than understanding the underlying compositional structure of a problem, RNNs rely on heuristics and exploit biases in the data.
Particularly relevant to the current work are the studies of \citet{lake2018generalization} and \citet{livska2018memorize}, who both present data sets specifically designed to reflect compositionality in their task accuracy.
Using their compositional tests, they show that vanilla seq2seq models do not readily generalize to solutions that exhibit an understanding of the underlying rule system of the tasks.



\subsection{Models}

Some recent approaches attack the lack of compositional behaviour of RNNs by designing models that have compositionality explicitly built in, for instance by equiping architectures with a series of specialized modules and a controller that composes them \citep[e.g.,][]{andreas2016modular,johnson2017clevr}. 
In this work, instead, we focus on inducing compositional solutions in RNN models, that are less rigid and generally require fewer supervision.

%

Our method draws inspiration from the work on compositional learning of \citet{hupkes2018learning}. 
The authors introduce the concept of \textit{Attentive Guidance}, a training signal given to the attention mechanism of a seq2seq model to induce more compositional solutions.
While they convincingly show that seq2seq models with attention can in fact implement such solutions (see \citet{baan2019realization} for an in-depth analysis), their model requires attention annotation of the training data, which may not always be available. 
In this work, we address this problem by designing a model that still aims to be compositional through the attention mechanism, but instead learns these patterns fully automatically, obtaining similar or even improved performance without the need of extra supervision.




Another line of work which exploits attention as a regularization technique is proposed by \citet{hudson2018compositional}, who introduce the Memory, Attention and Composition (MAC) cell.
The MAC cell consists of three components, whose communication within one cell is restricted to using attention.
An important limitation of the MAC cell is that the number of reasoning steps needs to be specified in advance. 
Our model, as vanilla seq2seq models, doesn't suffer from this limitation.


\begin{figure*}[!ht]
    \centering
    \includegraphics[width=0.9\linewidth]{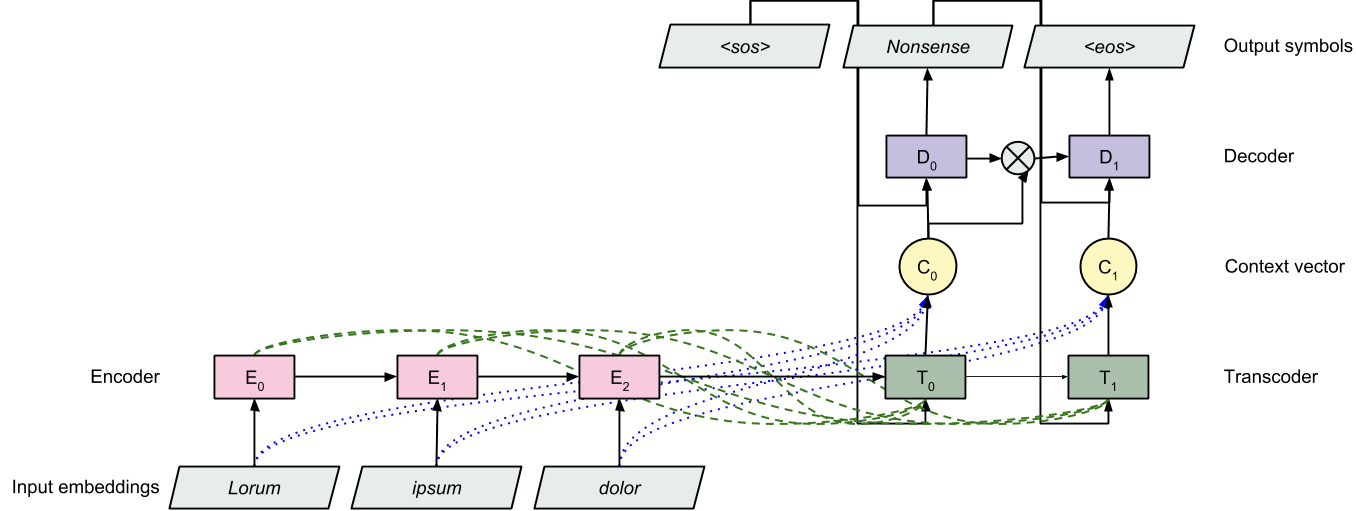}
    \caption{Schematic of the seq2attn architecture. The input sequence is processed by the encoder (E), after which the transcoder (T) generates context vectors, which are weighted means over the input embeddings.}
    \label{fig:architecture}
\end{figure*}

%% file: model.tex
\section{Model} 
\label{sec:model}


We propose \textit{seq2attn}, a novel attention-centric module that connects the encoder and decoder of a seq2seq model.\footnote{We will make our code available upon publication.}
The core component of seq2attn is the transcoder: a recurrent module that modulates the information flow between encoder and decoder by generating \textit{sparse} attention vectors using separate \textit{keys} and \textit{values}.
Below, we demonstrate and test how seq2attn can be used in combination with a vanilla encoder-decoder architecture.



\subsection{Encoder}
In our tests, we assume a standard recurrent encoder, that, given an input sequence $\{x_1, \dots, x_N\}$ and an embedding layer $\mathcal{E}^{enc}$, 
generates a sequence of outputs and hidden states:
\begin{gather}
\mathbf x^{enc}_{t} = \mathcal{E}^{enc}(x_{t}) \\
\mathbf y^{enc}_{t}, \mathbf h^{enc}_{t} = \mathcal{S}^{enc}(\mathbf x^{enc}_{t}, \mathbf h^{enc}_{t-1}) 
\label{eq:encoder}
\end{gather}
$\mathcal S$ is a recurrent state transition model, such as a vanilla RNN, LSTM or GRU.

\subsection{Transcoder}
The transcoder is initialized with $\mathbf h^{trans}_{0} = \mathbf h^{enc}_{N}$ and uses the hidden states of the encoder to compute context vectors $\mathbf c_t$ that will be passed to the decoder.

The input to the transcoder is the embedded output of te decoder (\cref{eq:decoder_output}):

\begin{gather}
    \mathbf x^{trans}_{t} = \mathcal{E}^{trans}(\hat{y}_{t-1}) \label{eq:trans_emb}\\
\mathbf y^{trans}_{t}, \mathbf h^{trans}_{t} = \mathcal{S}^{trans}(\mathbf x^{trans}_{t}, \mathbf h^{trans}_{t-1})
\label{eq:transcoder}
\end{gather}

The transcoder state is then used to query the hidden states of the encoder. The resulting scores are normalized using the Softmax function:
\begin{gather}
   \alpha_t(s) = \bm v_a^\top \cdot ReLU( \bm W_{a} \cdot [\mathbf h^{enc}_{t}; \mathbf h^{trans}_{t-1}] ) \displaybreak[0] \label{eq:attention-vector}\\
    \mathbf \pi_t(s) = \frac{\exp \alpha_t(s)} {\sum_{i=1}^N \exp \alpha_t(i)}\label{eq:softmax}
\end{gather}

Using the Softmax distribution often results in distributed vectors that attend to many input symbols at the same time, while an ideal compositional attention vector only focuses on the relevant parts of the input.
To force the transcoder to be more selective in the information it selects, we use Gumbel-Softmax, which allows us to draw from the categorical distribution computed in \cref{eq:softmax}, with continuous relaxation \citep{jang2016categorical, maddison2016concrete}.
The Straight-Through estimator is then used as a biased gradient estimator of the $\arg \max$ operator:
\begin{gather}
\tilde{\mathbf a}_t(s) = \frac{\exp \frac{\log \mathbf \pi_t(s) + g_s}{\tau} }{\sum_{i=1}^N \exp \frac{\log \mathbf \pi_t(i) + g_i}{\tau}}
\label{eq:gumbel}\\ 
\mathbf a_t = \tilde{\mathbf a}_t - \hat{\mathbf a}_{t} + \text{one\_hot}(\arg \max (\tilde{\mathbf a}_t))
\label{eq:st}
\end{gather}
The temperature $\tau$ can be interpreted as a measure of uncertainty. $\hat{\mathbf a}_{t}$ is a copy of $\tilde{\mathbf a}_{t}$ which we do not backpropagate through.
At inference time the stochasticity of Gumbel-Softmax is not needed, and $\arg \max$ is used as activation function.

The resulting attention weights are used to compute the context vectors that will be passed to the decoder:
\begin{gather}
\mathbf c_t = \sum_{i=1}^N \mathbf a_t(i) \cdot \mathbf x^{enc}_{i}
\label{eq:context-vector}
\end{gather}
Crucially, the context vector represents a weighted average of the \textit{input embeddings} of the encoder, while the weights $\mathbf a_t(i)$ are depending on the \textit{hidden states} of the encoder, thus introducing a separation between attention \textit{keys} and \textit{values} \citep[similar to, e.g.,][]{mino2017key, vaswani2017attention}.


\subsection{Decoder}\label{ssec:decoder}

The decoder of a seq2seq model is commonly initialized with the final hidden state of the encoder.
However, as this state vector encodes the entire input sequence, this type of initialization does not urge compositional behavior of the decoder.
When seq2attn is used, the decoder should be initialized with a fixed, \textit{learned} initialization vector.
In combination with using input embeddings as attention values (\cref{eq:context-vector}), this restricts the decoder to work only with disentangled representations of the input sequence, which encourages it to learn and process the individual meaning of all input symbols. 

To model outputs, the decoder uses the context vector $\mathbf c_t$, its own embedded output (identical to \cref{eq:trans_emb}) and a vector $\mathbf h_{t-1}^{dec}$ that integrates the current decoder hidden state with the context vector:
\begin{gather}
    \mathbf y^{dec}_{t}, \tilde{\mathbf h}^{dec}_{t} = \mathcal S^{dec} ( [\mathbf c_t;\mathcal{E}^{trans}(\hat{y}_{t-1})], {\mathbf h}^{dec}_{t-1} )
\label{eq:decoder}\\
\hat{y_t} = \arg \max (\text{Softmax}(\mathbf y_{t}^{dec}))\label{eq:decoder_output}
\end{gather}

Where $\mathbf h^{dec}_{t-1}$ is computed using an element-wise multiplication of the context vector with the previous hidden state of the decoder:
\begin{gather}
\mathbf h^{dec}_{t-1} = \tilde{ \mathbf h}^{dec}_{t-1} \odot \mathbf c_t
\label{eq:ff}
\end{gather}

This way of integrating the context vector with the decoder, which we call \textit{full focus}, makes the output of the decoder at decoding step $t$ more directly dependent on the current context vector $c_t$.

%% file: lookup_tables.tex
\section{Test Case 1: Lookup tables}
\label{sec:lookup_tables_experiments}

Our first test-case is the lookup table task introduced by \citet{livska2018memorize}.

\subsection{Task}
\begin{table}[t]
\centering
\resizebox{\linewidth}{!}{%
\begin{tabular}{l|lll}
 & \multicolumn{1}{c}{\begin{tabular}[c]{@{}c@{}}held-out\\ inputs\end{tabular}}
 & \multicolumn{1}{c}{\begin{tabular}[c]{@{}c@{}}held-out\\ compositions\end{tabular}} 
 & \multicolumn{1}{c}{\begin{tabular}[c]{@{}c@{}}held-out\\ tables\end{tabular}} \\ \hline
 Baseline & 38.25 $\pm$ 0.04 & 43.28 $\pm$ 0.09 & 7.86 $\pm$ 0.02 \\
Seq2attn & \textbf{100 $\pm$ 0.00} & \textbf{100 $\pm$ 0.00} & \textbf{100 $\pm$ 0.00}
\end{tabular}%
}
\caption{Average sequence accuracies and standard deviations of the baseline and seq2attn models on all lookup tables test sets.}
\label{tab:lookup-accuracies}
\end{table}
The core of the \emph{lookup table composition} domain consists in sequentially applying simple lookup table functions.
The functions to be applied are bijective mappings from the set of all \emph{n}-bit bitstrings onto itself.
Following \citet{livska2018memorize}, we focus on 3-bit strings, resulting in $2^3=8$ possible inputs and outputs.
We create 8 random table lookup functions, to which we refer with the names \texttt{t1}, \texttt{t2}, \ldots, \texttt{t8}.
Given the simplicity of the functions, the main challenge of the task resides in inferring that the input sequences should be treated \textit{compositionally}, rather than considered as a whole.

We borrow the setup presented in \citet{hupkes2018learning}, which differs slightly from the setup as it was originally presented.
In this setup, a typical input output example could be \texttt{001} \texttt{t1} \texttt{t2} $\rightarrow$ \texttt{001} \texttt{010} \texttt{111}.
Computing the output for this example requires the sequential application of \texttt{t1} to \texttt{001}, and then \texttt{t2} to the intermediate result.
Since two tables are to be applied in succession, we refer to such an examples as a \emph{binary composition}, as opposed to a \emph{unary composition} in which only one function has to be applied on the input.
The input bitstring and all intermediate outputs are included in the target output sequence.

\citet{livska2018memorize} train models on all 8 inputs for unary compositions and on 6 out of 8 input bitstrings of all binary compositions.
The remaining 2 \textit{held-out inputs} are used to test for generalization.
Following \citet{hupkes2018learning}, we do not include all 64 binary compositions in the training set, but leave out some for testing.
In particular, we create one test set that contains all binary compositions containing \texttt{t7} or \texttt{t8}, which are thus only seen in the training set as unary compositions.
We call this condition \textit{held-out tables}.
Of the remaining binary compositions, that contain only functions in $\{$\texttt{t1},~$\dots$,~\texttt{t6}$\}$, 8 randomly selected compositions are held out from the train set for all inputs, which form the \textit{held-out compositions} test set. 
Lastly, we remove 2 of the 8 inputs for each binary composition independently to form the \textit{held-out inputs} test set, which is similar to the generalization condition presented by \citet{livska2018memorize}.

\subsection{Results}
\begin{table}[t]
\centering
\resizebox{\linewidth}{!}{%
\begin{tabular}{l|p{0.6cm}p{1.4cm}p{0.0cm}p{0.6cm}p{1.4cm}p{0.0cm}p{0.6cm}l}
 & \multicolumn{2}{c}{\begin{tabular}[c]{@{}c@{}}held-out\\ inputs\end{tabular}}
 && \multicolumn{2}{c}{\begin{tabular}[c]{@{}c@{}}held-out\\ compositions\end{tabular}} 
 && \multicolumn{2}{c}{\begin{tabular}[c]{@{}c@{}}held-out\\ tables\end{tabular}} \\ \hline

Baseline+G & 34.17 & $\pm$ 8.25  && 38.54 & $\pm$ 12.39 && 8.16  & $\pm$ 3.57 \\
Baseline+E   & 82.50 & $\pm$ 12.42 && 85.42 & $\pm$ 12.39 && 31.08 & $\pm$ 7.85 \\
Baseline+F   & 85.83 & $\pm$ 16.50 && 91.67 & $\pm$ 11.79 && 30.03 & $\pm$ 16.12 \\
Baseline+T\vspace{1.2mm} & 43.33 & $\pm$ 12.30 && 47.40 & $\pm$ 15.33 && 3.99  & $\pm$ 2.70 \\

Baseline+GE  & 82.50 & $\pm$ 12.42 && 83.85 & $\pm$ 7.48  && 30.21 & $\pm$ 3.32 \\
Baseline+GF  & 69.17 & $\pm$ 21.25 && 76.04 & $\pm$ 13.28 && 4.69  & $\pm$ 1.47 \\
Baseline+GT  & 32.50 & $\pm$ 8.90  && 45.31 & $\pm$ 10.13 && 1.56  & $\pm$ 1.53 \\
Baseline+EF  & 85.00 & $\pm$ 9.35  && 82.29 & $\pm$ 18.46 && 24.13 & $\pm$ 2.99 \\
Baseline+ET  & \textbf{100.00} & \textbf{$\pm$ 0.00} && \textbf{100.00} & \textbf{$\pm$ 0.00} && 41.49 & $\pm$ 3.30 \\
Baseline+FT\vspace{1.2mm}  & 68.33 & $\pm$ 21.44 && 71.88 & $\pm$ 23.00 && 19.44 & $\pm$ 19.06 \\

Baseline+GEF & 74.17 & $\pm$ 36.53 && 72.40 & $\pm$ 37.94 && 37.33 & $\pm$ 22.10 \\
Baseline+GET & 97.50 & $\pm$ 3.54  && 98.44 & $\pm$ 1.28  && 24.31 & $\pm$ 17.87 \\
Baseline+GFT & 90.83 & $\pm$ 3.12  && 91.15 & $\pm$ 3.21  && 28.30 & $\pm$ 7.23 \\
Baseline+EFT\vspace{1.2mm} & 66.67 & $\pm$ 47.14 && 66.67 & $\pm$ 47.14 && 66.67 & $\pm$ 47.14 \\

Seq2attn     & \textbf{100.00} & \textbf{$\pm$ 0.00} && \textbf{100.00} & \textbf{$\pm$ 0.00} && \textbf{100.00} & \textbf{$\pm$ 0.00}
\end{tabular}%
}
\caption{Mean sequence accuracies and standard deviation on the lookup tables task of a baseline seq2seq model with additional components of seq2attn. \textbf{G}=Gumbel-Softmax, \textbf{E}=embeddings as attention values, \textbf{F}=full focus, \textbf{T}=transcoder.}
\label{tab:ablation}
\end{table}

We first compare the seq2attn architecture to a standard seq2seq model with an attention mechanism on generalization to new test examples.  
We establish the optimal parameters for both models using a grid search over a separate validation set.
Our search includes the type of RNN cell (\{GRU, LSTM\}), the embedding and RNN sizes (\{32, 64, 128, 256, 512, 1024\}) and the dropout rate (\{0, 0.2, 0.5\}).
The results are summarized in \cref{tab:hyperparams}.
The mini-batch size (1) and optimizer (Adam with default parameters \citep{kingma2014adam}) are fixed.
We train 10 models with the optimal parameters and report mean sequence accuracy.
For simplicity we will henceforth simply refer to this as the accuracy.

Our experiments confirm the findings previously presented by \citet{hupkes2018learning} and \citet{livska2018memorize}: Vanilla seq2seq models do not find generalizing solutions for the lookup table task (Table \ref{tab:lookup-accuracies}, first row).
Seq2attn, on the other hand, generalizes perfectly to data outside the training distribution. 
This first test confirms our hypothesized compositional bias of seq2attn.

\subsection{Ablation Study}
The difference between a traditional seq2seq and the seq2attn model can be summarized as the use of (i) a transcoder, (ii) the Gumbel-Softmax activation for the attention vector, (iii) using input embeddings as attention values and (iv) using full focus.
To assess the contributions of these components, we take the seq2attn model with optimal hyper-parameters as a base model, and increasingly ablate components.
The results of this study (\cref{tab:ablation}) indicate that, while some of the components of seq2attn cause an increase in accuracy on their own, no subset of them can match the performance of the full seq2attn model.

\subsection{Attention patterns}
\begin{figure*}[t]
    \centering
    \begin{subfigure}{.33\linewidth}
      \centering
      \includegraphics[width=0.80\linewidth,trim={2cm 0cm 1cm 0cm},clip]{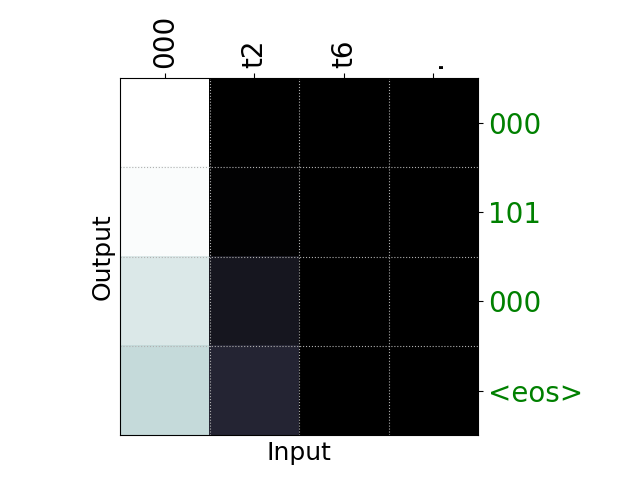}
      \caption{seq2seq}
      \label{fig:attn_baseline_lookup_heldout_input1}
    \end{subfigure}%
    \begin{subfigure}{.33\linewidth}
      \centering
      \includegraphics[width=0.80\linewidth,trim={2cm 0cm 1cm 0cm},clip]{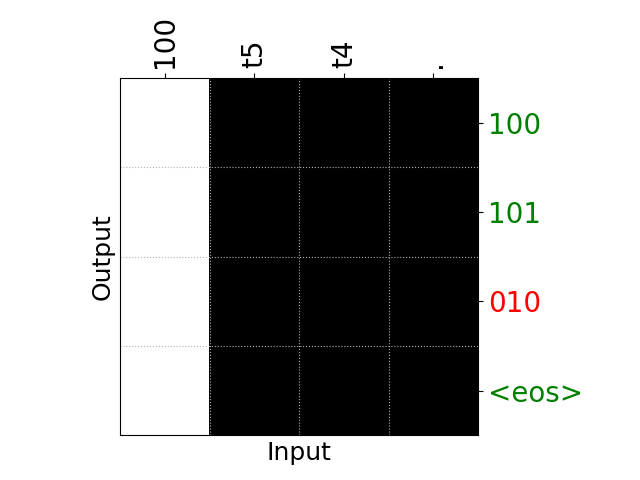}
      \caption{seq2seq}
      \label{fig:attn_baseline_lookup_heldout_input2}
    \end{subfigure}%
    \begin{subfigure}{.33\linewidth}
      \centering
      \includegraphics[width=0.80\linewidth,trim={2cm 0cm 1cm 0cm},clip]{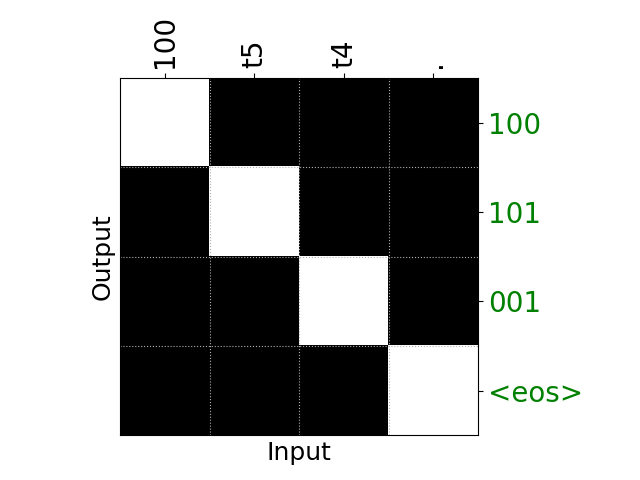}
      \caption{seq2attn}
      \label{fig:attn_seq2attn_lookup_heldout_input}
    \end{subfigure}
\caption{Examples of modeled attention patterns on held-out input examples of the lookup tables domain.}
\label{fig:attn_lookup}
\end{figure*}

As the modeled output of the decoder is highly dictated by the context vectors that it receives, we can gain insights into the types     of solutions the models are forming by studying their attention vectors.
As illustrated in Figure \ref{fig:attn_lookup}, seq2attn learns to generate a ``correct'' attention trace, attending to the right input at the right time.
Contrastingly, the baseline fails to capture a systematic pattern and produces a diffused attention instead or attends to irrelevant inputs, indicating that it does not utilize the attention mechanism to its full advantage.

\subsection{Overgeneralization}
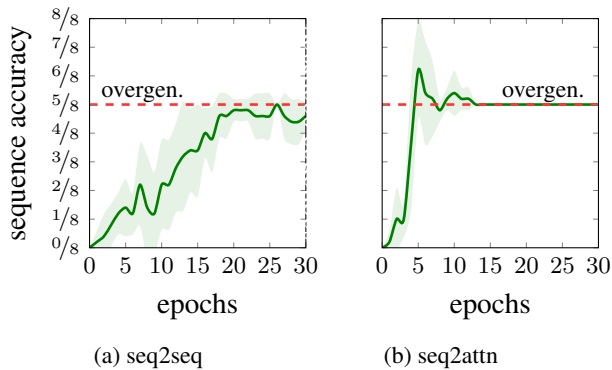
\begin{figure}[!h]
\centering
\begin{subfigure}{0.5\linewidth}
\input{Figures/corruption-lookup-baseline.tex}
\caption{seq2seq}
\end{subfigure}%
\begin{subfigure}{0.5\linewidth}
\input{Figures/corruption-lookup-seq2attn.tex}
\caption{seq2attn}
\end{subfigure}
\caption{Average accuracies on \emph{original targets} for the eight inputs in composition \texttt{t1 t2}. As three of these compositions are exceptions, we refer to accuracy higher than $\nicefrac{5}{8}$ as overgeneralization. The 95\% confidence interval is indicated.}
\label{fig:corruption-lookup}
\end{figure}

The results for the lookup tables task indicate that seq2attn performs much better than the baseline on data containing held-out inputs, tables and compositions. 
The model is thus better able to infer the compositional rules underlying the data.
To further explore seq2attn's bias towards compositionality, we test its behaviour when confronted with \textit{uncompositional} examples, that do not adhere to the previously mentioned rules.
Where a model unaware of the underlying task structure would have little problems learning such exceptions -- or in fact, would not realise that they are exceptions -- a model with a strong compositional bias may sometimes wrongly assume the exceptions also adhere to the underlying system, and \textit{overgeneralize} an inferred rule.
The extent to which a model overgeneralizes can thus be seen as a proxy for the strength of its compositional bias.
Whether overgeneralization is actually preferentiable behavior is depend on the task to be solved.

In the proposed setup, a small number of training instances are assigned adapted output targets.
We call these instances \emph{exceptions}.
The target output sequences of the exceptions are changed such that they can only be learned through memorization.
For the lookup table task, we adapt the training set such that one composition, \texttt{t1 t2}, is an exception to the general rules for three out of the eight existing input bitstrings.
In the target output, the third bitstring is replaced with a randomly selected bitstring, thus changing the application of table \texttt{t2} in this context.
Both the three adapted samples and the other five unadapted samples for \texttt{t1 t2} are included in the training set.

While training a model, we monitor the output sequences generated for these exceptions.
The accuracy on the original targets is reported to identify whether the model is processing the exceptions compositionally despite being exposed to the adapted targets in the training set, i.e., whether the model is overgeneralizing.

Figure~\ref{fig:corruption-lookup} displays the accuracy on the \emph{original} targets of all eight inputs in composition \texttt{t1 t2} over the first 30 training epochs.
While both the baseline and seq2attn learn to memorize the three exceptions, only seq2attn shows a strong bias to treat the inputs compositionally before memorizing the adapted targets.
The performance goes as high up as $\nicefrac{8}{8}$ between the fifth and fifteenth epoch for differently initialized models, before dropping to $\nicefrac{5}{8}$.
This indicates that the rules are learned before the adapted instances are memorized as exceptions to these rules.

%% file: Figures/corruption-lookup-baseline.tex
\begin{tikzpicture}
\begin{axis}[
    smooth,
    anchor=above north,
    width=1.15\linewidth,height=1.2\linewidth,
    ymax=1,
    xmax=30,
    xmin=0,
    xtick={0, 5, 10, 15, 20, 25, 30, 35, 40, 45, 50},
    tickwidth=2pt,
    ylabel=sequence accuracy,
    xlabel=epochs,
    ytick={0,0.125,0.25,0.375,0.5,0.6250,0.75,0.8750,1.0},
    yticklabels={${\nicefrac{0}{8}}$,$\nicefrac{1}{8}$,$\nicefrac{2}{8}$,$\nicefrac{3}{8}$,$\nicefrac{4}{8}$,$\nicefrac{5}{8}$,$\nicefrac{6}{8}$,$\nicefrac{7}{8}$,$\nicefrac{8}{8}$ },
    axis on top,
    ticklabel style = {font=\small}
    ]

        \addplot [name path=upper,draw=none] table[x=epochs, y expr=\thisrow{average}+\thisrow{confidence}] {Data/corruption-lookup-baseline.csv};
        \addplot [name path=lower,draw=none] table[x=epochs, y expr=\thisrow{average}-\thisrow{confidence}] {Data/corruption-lookup-baseline.csv};
        \addplot [fill=green_mtplotlib!10] fill between[of=upper and lower];
        \addplot [mark=none, color=green_mtplotlib, line width=1pt] table [x=epochs,y=average] {Data/corruption-lookup-baseline.csv};
        \node at (axis cs:14.5,0.75) [anchor=north east, font=\small] {overgen.};
        \addplot [color=red!80, line width=1pt, style=dashed] coordinates{(0,0.625) (30,0.625)};
\end{axis}
\end{tikzpicture}

%% file: Figures/corruption-lookup-seq2attn.tex
\begin{tikzpicture}
\begin{axis}[
    smooth,
    width=1.15\linewidth,height=1.2\linewidth,
    anchor=above north,
    ymax=1,
    xmax=30,
    xmin=0,
    xtick={0, 5, 10, 15, 20, 25, 30, 35, 40, 45, 50},
    tickwidth=2pt,
    xlabel=epochs,
    ylabel={\textcolor{white}{sequence accuracy}},
    ytick={0,0.125,0.25,0.375,0.5,0.6250,0.75,0.8750,1.0},
    yticklabels={$\textcolor{white}{\nicefrac{0}{8}}$,$\textcolor{white}{\nicefrac{1}{8}}$,$\textcolor{white}{\nicefrac{2}{8}}$,$\textcolor{white}{\nicefrac{3}{8}}$,$\textcolor{white}{\nicefrac{4}{8}}$,$\textcolor{white}{\nicefrac{5}{8}}$,$\textcolor{white}{\nicefrac{6}{8}}$,$\textcolor{white}{\nicefrac{7}{8}}$,$\textcolor{white}{\nicefrac{8}{8}}$ },
    axis on top,
    ticklabel style = {font=\small}
    ]

        \addplot [name path=upper,draw=none] table[x=epochs, y expr=\thisrow{average}+\thisrow{confidence}] {Data/corruption-lookup-seq2attn.csv};
        \addplot [name path=lower,draw=none] table[x=epochs, y expr=\thisrow{average}-\thisrow{confidence}] {Data/corruption-lookup-seq2attn.csv};
        \addplot [fill=green_mtplotlib!10] fill between[of=upper and lower];
        \addplot [mark=none, color=green_mtplotlib, line width=1pt] table [x=epochs,y=average] {Data/corruption-lookup-seq2attn.csv};
        \node at (axis cs:29.5,0.75) [anchor=north east, font=\small] {overgen.};
        \addplot [color=red!80, line width=1pt, style=dashed] coordinates{(0,0.625) (30,0.625)};
\end{axis}
\end{tikzpicture}

%% file: scan.tex
\section{Test Case 2: SCAN}
\begin{table*}[t]
    \centering
\begin{tabular}{l|ll}
              & Baseline               & Seq2attn              \\ \hline
Lookup tables & 128, 512, 1, GRU, 0.5  & 256, 256, 1, GRU, 0.5 \\
SCAN          & 200, 200, 2, LSTM, 0.5 & 512, 512, 1, GRU, 0.5
\end{tabular}
\caption{Hyperparameters (embedding dimensions, RNN dimensions, RNN layers, RNN type, dropout rate) used for both the seq2seq baseline and seq2attn model for both tasks.}
\label{tab:hyperparams}
\end{table*}
\label{sec:scan_experiments}
While the lookup table task provides an excellent test case to evaluate the compositional abilities of a neural network model, its simplicity limits the conclusions that can be drawn about the usability of seq2attn in more challenging domains.
In this section, we evaluate seq2attn on \textit{SCAN} \citep{lake2018generalization}, a task involving mapping navigational commands to sequences of output actions.

\subsection{Task}
The input commands of the SCAN task are composed of a small set of predefined atomic commands (\textit{jump}, \textit{walk}, \textit{run} and \textit{look}), modifiers (\textit{twice}, \textit{thrice}, \textit{around}, \textit{opposite}, \textit{left} and \textit{right}) and conjunctions (\textit{after} and \textit{and}) that are combined via a limited context free grammar, such that there are no ambiguities.
An example input is \textit{jump after walk left twice}, where the learning agent has to mentally perform these actions in a 2-dimensional grid and output the sequence of actions it takes: ``I\_TURN\_LEFT \hspace{.1em} I\_WALK \hspace{.1em} I\_TURN\_LEFT \hspace{.1em} I\_WALK \hspace{.1em} I\_JUMP''.
For full details of the data set and experiments, we refer to \citet{lake2018generalization}.

\citeauthor{lake2018generalization} use three different train-test distributions of the total of 20.910 examples.
They show that vanilla seq2seq models are able to almost perfectly generalize when the data is randomly split in a training and testing set, but that they are unfit for generalizing to longer test sequences and for one-shot learning to commands seen only in their atomic form.
Later, \citet{loula2018rearranging} proposed a new set of experiments based on the same task, which they argue are better suited for assessing systematic compositionality.
We focus on experiments 2 and 3 of their paper.

Experiment 2 contains four different train-test distributions as there are four primitive commands involved.
For all four conditions, the test set is the same.
This test set consists of all examples that contain ``jump around right'' in their input sequences.
The first condition, which is called \textbf{0 fillers}, contains no subsequences of the form ``\textit{primitive} around right'' in the training set, where \textit{primitive} is either of the four primitives ``jump'', ``look'', ``run'' or ``walk''.
This condition should thus test whether a model can induce a compositional understanding of ``jump around right'' by showing those symbols (``jump'', ``around'' and ``right'') only in different contexts.
The next three conditions, \textbf{1 filler}, \textbf{2 fillers} and \textbf{3 fillers}, are considered increasingly easier.
They retain the same test set, but increasingly add more examples to the train set of the template ``\textit{primitive} around right''.
\textbf{1 filler} adds all examples of this template where \textit{primitive} is ``look''.
\textbf{2 fillers} and \textbf{3 fillers} add ``walk'' and ``run'' respectively.

As \citet{loula2018rearranging} observed a great difference in performance between the \textbf{0 fillers} and \textbf{1 filler} conditions, they zoom in on these conditions in experiment 3.
The \textbf{0 fillers} condition contains 0 examples with the subsequence ``\textit{primitive} around right'' in the training set.
The \textbf{1 filler} condition contains 1.100 of those, namely all examples which contain the subsequence ``look around right''.
In experiment 3, they test a more smooth and dense transition from the \textbf{0 fillers} condition to the \textbf{1 filler} conditions.
They accomplish this by taking the training set of the \textbf{0 fillers} condition and adding respectively 1, 2, 4, 8, 16, 32, 64, 128, 256, 512 and 1024 extra examples containing the subsequence ``look around right'', resulting in 11 new training sets.
The test set is again the same as in experiment 2.

\subsection{Results}
\begin{figure*}[!h]
\centering
    \begin{subfigure}{.5\linewidth}
      \centering
      \input{Figures/scan_exp2}
        \vspace{-3mm}
    \end{subfigure}%
    \begin{subfigure}{.5\linewidth}
      \centering
      \input{Figures/scan_exp3}
    \end{subfigure}
    \caption{Mean sequence accuracies on experiment 2 (left) and 3 (right) of \citet{loula2018rearranging}. The bootstrapped 95\% confidence intervals are indicated with error bars.}
\label{fig:scan-results}
\end{figure*}
We compare a baseline seq2seq to the seq2attn architecture on these two tasks.
First, we perform a grid search using a random split of the data to find the optimal parameters for seq2attn.
The results of this are summarized in \cref{tab:hyperparams}.
As a baseline, we used the model which \citet{lake2018generalization} found to be overall best performing, which is a seq2seq model with 2-layer LSTMs, 200 hidden units per layer and a dropout rate of 0.5.
For comparison to the seq2attn model, we also added an attention mechanism, which was missing in the original model.
For all reported results we ran these models 10 times with random weight initialization.
Since experiments 2 and 3 by \citet{loula2018rearranging} do not have validation sets for early stopping, we ran all models for 50 epochs.

Firstly, we confirm the findings of \citet{lake2018generalization} and \citet{loula2018rearranging} (see \cref{fig:scan-exp2}).
A vanilla seq2seq with attention is able to perform analogical generalization (95.19\% accuracy): it requires examples of \textbf{1 filler} only to generalize to other fillers of the same template. 
On the other hand, it is not able to apply ``right'' and ``around'' to a primitive verb in a productive way, when they were never seen together (0.26\% accuracy, \textbf{0 fillers} condition). 
When we look at seq2attn, we notice how not only it is able to perform analogical generalization (94.32\% accuracy, \textbf{1 filler}) but, to a certain extent, it is also able to generalize productively in the \textbf{0 fillers} condition (36.23\% accuracy).

In Figure \ref{fig:scan-exp3} we report the results for experiment 3 of \citet{loula2018rearranging} where we consider the \textbf{0 fillers} condition of Experiment 2 and progressively add extra training examples from \textbf{1 filler}. 
As \citet{loula2018rearranging} observed, performance of a seq2seq model ramps up as more samples are injected in the training set. 
Yet, the fact that performance increases gradually and takes long to peak (at 512 examples) suggests that rather than systematically understanding the rule, the model is piling up evidence for a very specific pattern. 
The situation is quite different for the seq2attn model, whose performance spikes much earlier, reaching a plateau at 16 examples already. 
Interestingly, the performance peak is also at 512, but with an improvement of just over 5 percentage points over 16 examples vs. 
approximately 50 percentage points improvement in the case of baseline. 
Seq2attn seems then to show evidence for an opposite interpretation, namely for a network that, to a certain extent, is able to induce the compositional rules. 
A property that is often linked to sample efficiency \citep{lake2018generalization}.

\begin{figure*}[!ht]
\centering
\begin{subfigure}{.33\linewidth}
  \centering
  \includegraphics[width=0.77\linewidth,trim={3cm 0cm 0.9cm 0cm},clip]{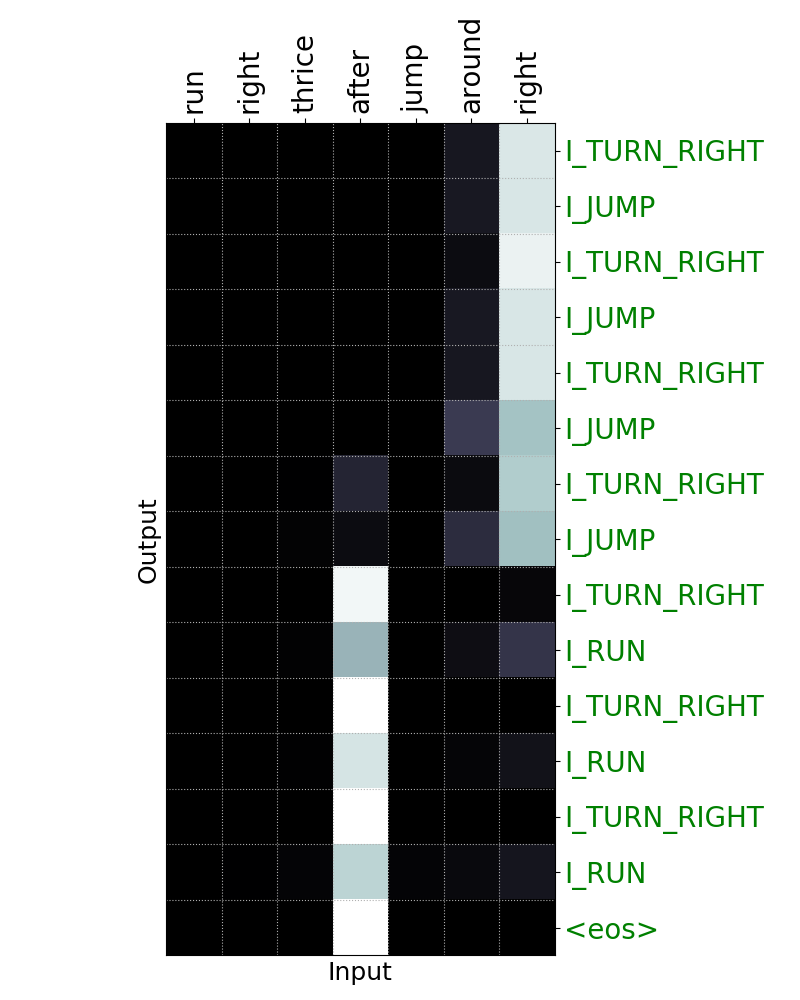}
  \caption{seq2seq}
  \label{fig:scan_attn_baseline}
\end{subfigure}%
\begin{subfigure}{.33\linewidth}
  \centering
  \includegraphics[width=0.77\linewidth,trim={3cm 0cm 0.9cm 0cm},clip]{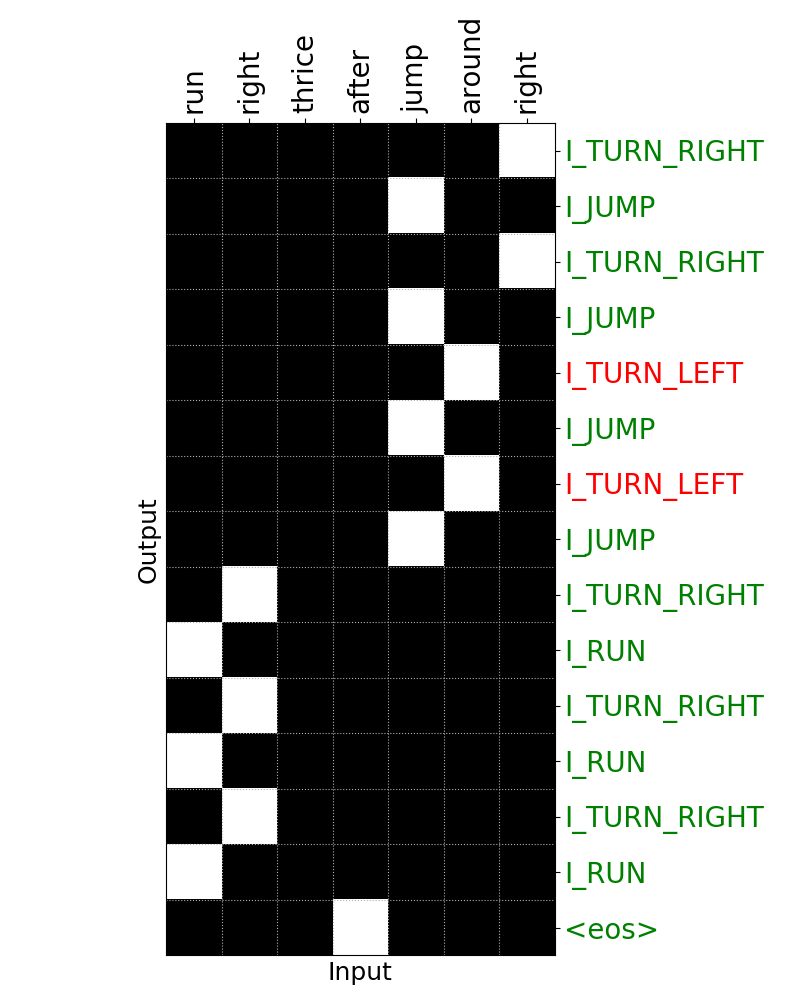}
  \caption{seq2attn}
  \label{fig:scan_attn_seq2attn_incorrect}
\end{subfigure}%
\begin{subfigure}{.33\linewidth}
  \centering
  \includegraphics[width=0.77\linewidth,trim={3cm 0cm 0.9cm 0cm},clip]{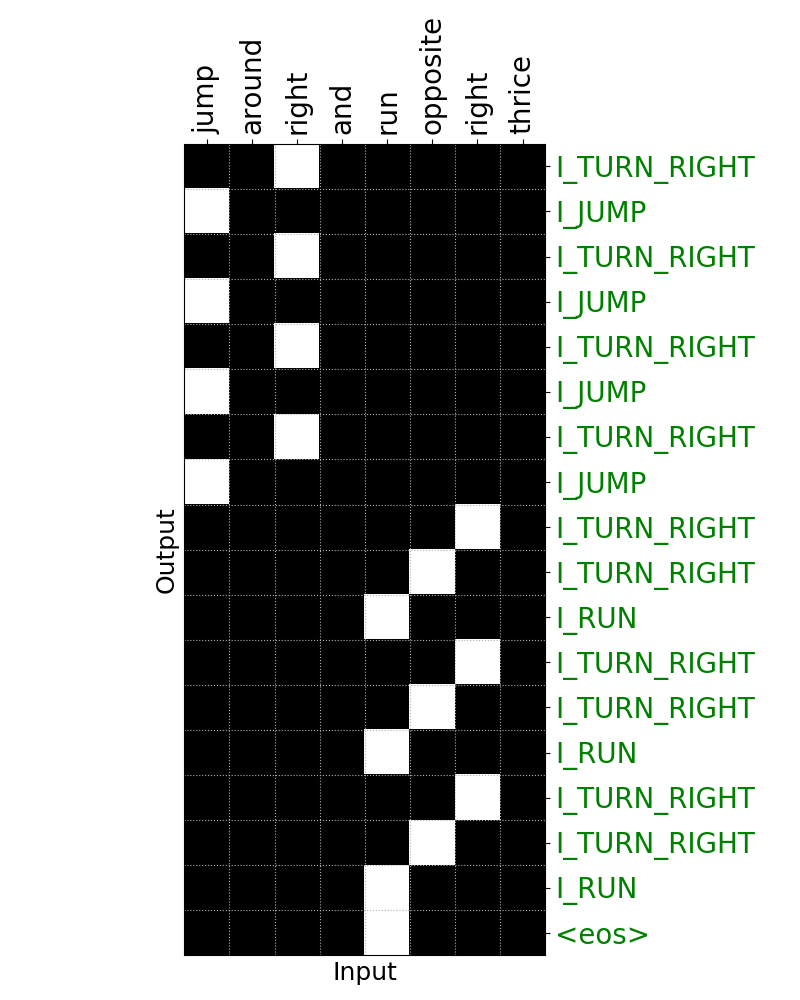}
  \caption{seq2attn}
  \label{fig:scan_attn_seq2attn_correct}
\end{subfigure}
\caption{Examples of attention patterns on the \textbf{0 fillers} condition. Seq2attn models the output incorrectly when the attention pattern is incorrect.}
\label{fig:scan_attn}
\end{figure*}

\subsection{Analysis}
In Figure \ref{fig:scan_attn}, we now look at some attention patterns for the \textbf{0 fillers} condition.
While baseline models emit sparser and more informative attentional patterns here than in the lookup table task, they still are locally diffused and, more importantly, do not maintain a systematic input-output alignment, which suggests that the models are not understanding the rules of the task, but use a pattern matching strategy instead.
On the contrary, seq2attn shows always fully sparse, one-hot attention patterns.
\Cref{fig:scan_attn_seq2attn_correct} shows how the model usually aligns outputs to their respective primitive commands or directions in the input sequence, e.g., ``I\_JUMP'' aligns to ``jump'', and ``I\_TURN\_RIGHT'' aligns to ``right''.
A modifier like ``opposite'' is used as an indicator to repeat the last modeled directional action.

Seq2attn reaches an accuracy of 36.23\% on the \textbf{0 fillers} condition, which still leaves room for improvement.
However, the attentional patterns quickly show the main cause of error.
\Cref{fig:scan_attn_seq2attn_incorrect} shows how the model outputs ``I\_TURN\_LEFT'' instead of ``I\_TURN\_RIGHT'' whenever it attends to the input ``around''.
Whenever the model does attend to ``right'', as is the expected, optimal behavior, the output is correct.
This behavior can be easily explained by analyzing the data that the model was trained on.
The input ``around'' has only been encountered within the context ``\emph{primitive} around left'' during training.
Thus, within this context, ``around'' and ``left'' could be used synonymically by the transcoder to communicate to the decoder to output ``I\_TURN\_LEFT''.
The great majority of errors on this task by seq2attn have the same cause.
Although seq2attn still does not perfectly solve the task, contrary to a standard seq2seq model, it provides an immediate understanding of the root of this.

\subsection{Overgeneralization}

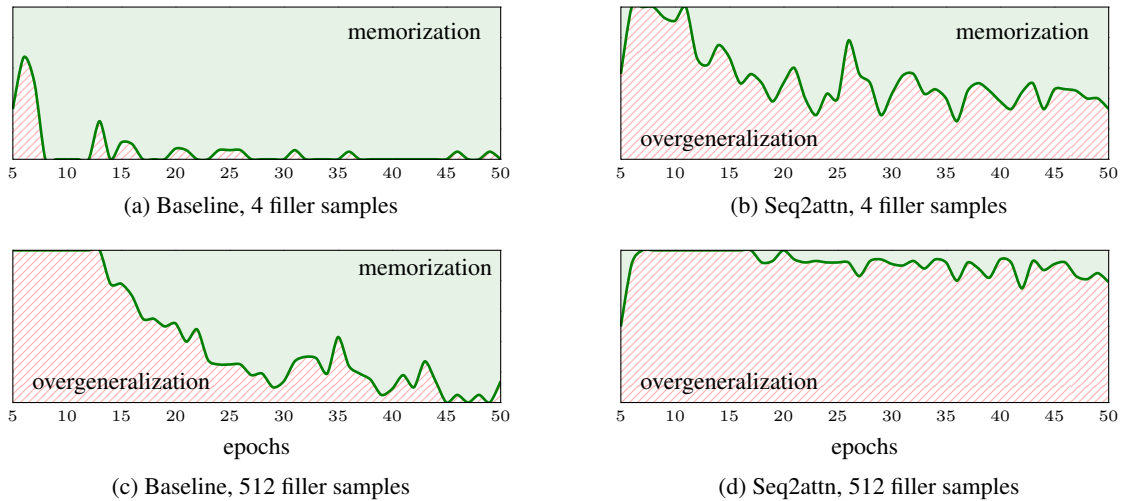
\begin{figure*}[!ht]
\centering
\begin{subfigure}{0.5\linewidth}
\centering
\input{Figures/corruption-scan-baseline/corruption-scan-baseline-4.tex}
\vspace{-1mm}
\caption{Baseline, 4 filler samples}
\vspace{0.25cm}
\end{subfigure}%
\begin{subfigure}{0.5\linewidth}
\centering
\input{Figures/corruption-scan-seq2attn/corruption-scan-seq2attn-4.tex}
\vspace{-1mm}
\caption{Seq2attn, 4 filler samples}
\vspace{0.25cm}
\end{subfigure}
\begin{subfigure}{0.5\linewidth}
\centering
\input{Figures/corruption-scan-baseline/corruption-scan-baseline-512.tex}
\vspace{-1mm}
\caption{Baseline, 512 filler samples}
\end{subfigure}%
\begin{subfigure}{0.5\linewidth}
\centering
\input{Figures/corruption-scan-seq2attn/corruption-scan-seq2attn-512.tex}
\vspace{-1mm}
\caption{Seq2attn, 512 filler samples}
\end{subfigure}
\caption{Mean sequence accuracy on the \emph{original target} of ``jump around right'' of multiple models as training progresses. The distribution was normalized for cases in which the output emitted was neither the original or the adapted target.}
\label{fig:corruption-scan}
\end{figure*}

To assess seq2attn's overgeneralization abilities for the SCAN task, we repeated experiment 3.
In addition to gradually adding samples indicating the correct interpretation of ``\emph{primitive} around right'', we also added a single exception for ``jump around right'' to the training set. 
The target for this sequence, originally consisting of four repetitions of ``I\_TURN\_RIGHT \hspace{.1em} I\_JUMP'', was modified to consist of only two repetitions.


For all conditions of experiment 3, we added the exception to the training set, trained multiple randomly initialized models, and monitored the output sequences generated for this exception over the course of training. 
In Figure~\ref{fig:corruption-scan}, we visualize the distribution over the adapted and original targets for the conditions with 4 and 512 filler samples respectively.
Note that the models have implicit and explicit evidence for the correct application of the rules for ``\emph{primitive} around right'': explicit evidence through training examples containing ``look around right`` subsequences, and implicit evidence through training samples including ``around'' or ``right'' seen in different contexts.

Both models exhibit overgeneralization behavior for SCAN. Generally, overgeneralization occurs at the start of the training process and precedes memorization of the adapted target. However, the baseline model needs a substantially larger amount of explicit evidence to overgeneralize as much as the seq2attn model. The condition where 512 filler samples are included illustrates that the tendency to overgeneralize does not necessarily relate to the overall task performance. For this condition, seq2attn and the baseline yield similar sequence accuracies in the original setup of experiment 3 (see Figure~\ref{fig:corruption-scan}), but seq2attn overgeneralizes more frequently, indicating that seq2attn has a stronger compositional bias. 


%% file: Figures/scan_exp2.tex
\begin{tikzpicture}
\begin{axis}[
    ybar,
    axis x line*=bottom,
    axis y line*=left,
    symbolic x coords={0,1,2,3},
    anchor=above north,
    ymax=110,
    tickwidth=2pt,
    bar width=13pt,
    enlarge x limits = 0.2,
    xtick=data,
    xticklabel style={align=center},
    ylabel=sequence accuracy,
    xlabel=Number of primitive fillers used for training,
    legend columns=2,
    legend style={at={(1.05,1.15)}},
    height=0.62\linewidth,
    width=0.9\linewidth
    ]
        \addplot [style={fill=green_mtplotlib},error bars/.cd, y dir=both,y explicit]  
        table [x=test set, y=mean,y error=conf, col sep=comma,] {./Data/scan_exp2_seq2attn.csv};	

        \addplot [pattern=north east lines, pattern color=red,error bars/.cd, y dir=both,y explicit]  
        table [x=test set, y=mean,y error=conf, col sep=comma,] {./Data/scan_exp2_baseline.csv};
        
        \legend{seq2attn,baseline}	
\end{axis}
\end{tikzpicture}

%% file: Figures/scan_exp3.tex
\begin{tikzpicture}
\begin{axis}[
    ybar,
    axis x line*=bottom,
    axis y line*=left,
    symbolic x coords={1,2,4,8,16,32,64,128,256,512,1024},
    anchor=above north,
    ymax=110,
    tickwidth=2pt,
    bar width=4pt,
    enlarge x limits = 0.06,
    xtick=data,
    xticklabel style={align=center, font=\small},
    ylabel=sequence accuracy,
    xlabel=Number of examples used for training,
    height=0.75\linewidth,
    width=0.9\linewidth
    ]
        \addplot [style={fill=green_mtplotlib},error bars/.cd, y dir=both,y explicit]  
        table [x=test set, y=mean,y error=conf, col sep=comma,] {./Data/scan_exp3_seq2attn.csv};	

        \addplot [pattern=north east lines, pattern color=red,error bars/.cd, y dir=both,y explicit]  
        table [x=test set, y=mean,y error=conf, col sep=comma,] {./Data/scan_exp3_baseline.csv};
        
\end{axis}
\end{tikzpicture}

%% file: Figures/corruption-scan-baseline/corruption-scan-baseline-4.tex
\small
\begin{tikzpicture}
\begin{axis}[
    smooth,
    width=1.\linewidth, height=.45\linewidth,
    anchor=above north,
    ymax=1,
    xmax=50,
    xmin=5,
    xtick={0, 5, 10, 15, 20, 25, 30, 35, 40, 45, 50},
    ytick={},
    yticklabels={,,},
    tickwidth=0pt,
    axis on top,
    tick label style={font=\tiny},
    legend style={draw=none, fill opacity=0.0, draw opacity=1, text opacity=1}
    ]

        \addplot [name path=upper, draw=none, forget plot] table[x=epochs, y=upper] {Data/exception_learning_scan_baseline.csv};
        \addplot [name path=uncorrupted, mark=none, color=green_mtplotlib, line width=1pt, forget plot] table [x=epochs,y=4] {Data/exception_learning_scan_baseline.csv};
        \addplot [name path=lower, draw=none, forget plot] coordinates{(0,0) (50,0)};
        \addplot [fill=green_mtplotlib!10, area legend] fill between[of=upper and uncorrupted];
        \addplot [pattern=north east lines, pattern color=red!30, area legend] fill between[of=lower and uncorrupted];
        \node at (axis cs:49.0,0.95) [anchor=north east, font=\small] {memorization};
\end{axis}
\end{tikzpicture}

%% file: Figures/corruption-scan-seq2attn/corruption-scan-seq2attn-4.tex
\small
\begin{tikzpicture}
\begin{axis}[
    smooth,
    width=1.0\linewidth, height=0.45\linewidth,
    anchor=above north,
    ymax=1,
    xmax=50,
    xmin=5,
    xtick={0, 5, 10, 15, 20, 25, 30, 35, 40, 45, 50},
    ytick={},
    tickwidth=0pt,
    tick label style={font=\tiny},
    axis on top,
    yticklabels={,,}
    ]

        \addplot [name path=upper, draw=none, forget plot] table[x=epochs, y=upper] {Data/exception_learning_scan_seq2attn.csv};
        \addplot [name path=uncorrupted, mark=none, color=green_mtplotlib, line width=1pt, forget plot] table [x=epochs,y=4] {Data/exception_learning_scan_seq2attn.csv};
        \addplot [name path=lower, draw=none, forget plot] coordinates{(0,0) (50,0)};
        \addplot [fill=green_mtplotlib!10, area legend] fill between[of=upper and uncorrupted];
        \addplot [pattern=north east lines, pattern color=red!30, area legend] fill between[of=lower and uncorrupted];
        \node at (axis cs:24.0,0.25) [anchor=north east, font=\small] {overgeneralization};
        \node at (axis cs:49.0,0.95) [anchor=north east, font=\small] {memorization};
\end{axis}
\end{tikzpicture}

%% file: Figures/corruption-scan-baseline/corruption-scan-baseline-512.tex
\small
\begin{tikzpicture}
\begin{axis}[
    smooth,
    width=1.0\linewidth, height=0.45\linewidth,
    anchor=above north,
    ymax=1,
    xmax=50,
    xmin=5,
    xtick={0, 5, 10, 15, 20, 25, 30, 35, 40, 45, 50},
    ytick={},
    tickwidth=0pt,
    xlabel=epochs,
    tick label style={font=\tiny},
    axis on top,
    yticklabels={,,}
    ]

        \addplot [name path=upper, draw=none, forget plot] table[x=epochs, y=upper] {Data/exception_learning_scan_baseline.csv};
        \addplot [name path=uncorrupted, mark=none, color=green_mtplotlib, line width=1pt, forget plot] table [x=epochs,y=512] {Data/exception_learning_scan_baseline.csv};
        \addplot [name path=lower, draw=none, forget plot] coordinates{(0,0) (50,0)};
        \addplot [fill=green_mtplotlib!10, area legend] fill between[of=upper and uncorrupted];
        \addplot [pattern=north east lines, pattern color=red!30, area legend] fill between[of=lower and uncorrupted];
        \node at (axis cs:24.0,0.25) [anchor=north east, font=\small] {overgeneralization};
        \node at (axis cs:50,1.00) [anchor=north east, font=\small] {memorization};

\end{axis}
\end{tikzpicture}

%% file: Figures/corruption-scan-seq2attn/corruption-scan-seq2attn-512.tex
\small
\begin{tikzpicture}
\begin{axis}[
    smooth,
    width=1.0\linewidth, height=0.45\linewidth,
    anchor=above north,
    ymax=1,
    xmax=50,
    xmin=5,
    xtick={0, 5, 10, 15, 20, 25, 30, 35, 40, 45, 50},
    ytick={0, 0.5, 1.0},
    tickwidth=0pt,
    xlabel=epochs,
    tick label style={font=\tiny},
    axis on top,
    yticklabels={,,}
    ]

        \addplot [name path=upper, draw=none, forget plot] table[x=epochs, y=upper] {Data/exception_learning_scan_seq2attn.csv};
        \addplot [name path=uncorrupted, mark=none, color=green_mtplotlib, line width=1pt, forget plot] table [x=epochs,y=512] {Data/exception_learning_scan_seq2attn.csv};
        \addplot [name path=lower, draw=none, forget plot] coordinates{(0,0) (50,0)};
        \addplot [fill=green_mtplotlib!10, area legend] fill between[of=upper and uncorrupted];
        \addplot [pattern=north east lines, pattern color=red!30, area legend] fill between[of=lower and uncorrupted];
        \node at (axis cs:24.0,0.25) [anchor=north east, font=\small] {overgeneralization};
\end{axis}
\end{tikzpicture}

%% file: conclusion.tex
\section{Discussion}
\label{sec:discussion}
In search for a neural network architecture that exhibits a bias towards systematic generalization, we introduced \textbf{seq2attn}, a recurrent attention-centric module that controls the information flow from encoder to decoder.
We installed this module in a standard recurrent encoder-decoder architecture. 
To quantify its capabilities in terms of systematic compositionality, we tested the model on the lookup table and SCAN tasks.
%

On both tasks, we see significant improvements compared to a standard recurrent seq2seq model, providing evidence for a compositional bias in the system.
Furthermore, because the architecture relies heavily on its attention mechanism, its solutions can more easily be interpreted by looking at the generated attention patterns.
This provides opportunities for analyzing what the model has learned as well as for detecting potential biases in the training set.

Although on the considered tasks, which are specifically designed to evaluate compositionality, seq2attn leads to clear improvements, its contribution could not have been observed when considering a task for which the test accuracy is not directly linked to compositionality, such as natural language modeling and translation.
We argue that, for those cases, additional assessment methods are needed to compare the compositional skills of different models.
We propose one such method, which involves monitoring to what extent a model \emph{overgeneralizes}.
We show how a model with seq2attn, for both tasks, has a greater tendency to overgeneralize than the baseline.

A possible limitation of the design of seq2attn is that the flow of information from transcoder to decoder is very rigid.
Possible solutions could be found in the use of less skewed activations than the Gumbel-Softmax such as Sparsemax \citep{martins2016softmax}, or allowing the transcoder to communicate multiple embeddings using adaptive computation time \citep{graves2016adaptive}.

Importantly, seq2attn is not tied to a particular type of seq2seq architecture.
In future work, we plan to install it into other popular seq2seq architectures such as convolutional seq2seq \citep{gehring2017convolutional} and Transformer models  \citep{vaswani2017attention}. 